\documentclass[11pt]{article}

\usepackage[final]{acl}

\usepackage{times}
\usepackage{latexsym}

\usepackage[T1]{fontenc}

\usepackage[utf8]{inputenc}

\usepackage{microtype}

\usepackage{inconsolata}

\usepackage{graphicx}

\usepackage{amsmath}
\usepackage{booktabs}
\usepackage{amssymb}
\usepackage{multirow}
\usepackage{pifont}

\title{Hybrid Autoregressive-Diffusion Model for Real-Time Sign Language Production}

\author{Maoxiao Ye \quad Xinfeng Ye \quad Mano Manoharan \\
  University of Auckland, New Zealand \\
  \texttt{\{xye804@aucklanduni.ac.nz, x.ye@auckland.ac.nz, mano.manoharan@auckland.ac.nz\}}}

\begin{document}
\maketitle
\begin{abstract}

Earlier Sign Language Production (SLP) models typically relied on autoregressive decoding, which naturally preserves temporal causality but suffers from error accumulation at inference time. More recent diffusion-based approaches improve generation quality through iterative denoising, yet their sequence-level refinement process introduces substantial latency. To address this trade-off, we propose \textbf{HybridSign}, a hybrid autoregressive-diffusion model for low-latency sign language production that combines causal frame generation with flow-based diffusion refinement. A \textbf{Multi-Scale Pose Representation} module captures fine-grained articulator features, while a \textbf{Confidence-Aware Causal Attention} mechanism leverages joint-level confidence scores to improve robustness under noisy 2D pose observations. Experiments on PHOENIX14T and How2Sign show that HybridSign consistently achieves the best quality--efficiency trade-off among the compared baselines. On the How2Sign test split, it reaches BLEU-1/4 scores of 30.12/6.48 and DTW of 3.89, while reducing time-to-first-frame to 5.90\,s and increasing throughput to 10.17 FPS under a 60-frame evaluation protocol.

\end{abstract}

\section{Introduction}

Sign language relies on coordinated body, hand, and facial motion, making automatic sign language production a challenging structured generation problem. A practical SLP system must preserve temporal coherence, articulate fine-grained local motion, and respond with low latency. Existing approaches typically favor either autoregressive models~\cite{atmc, ptslp, gen-obt}, which model temporal dependencies well but suffer from exposure bias during inference, or diffusion models~\cite{Huang2021TowardsFA, g2pddm, gcdm}, which generate higher-quality poses but incur substantial sampling cost.

We propose \textbf{HybridSign}, a hybrid autoregressive-diffusion model that combines the strengths of both paradigms for low-latency SLP. In this paper, low latency refers to the ability to emit the first pose frame quickly and then continue generation frame by frame, which is more informative for interactive use than reporting only the total offline generation time. The autoregressive pathway provides causal frame generation, while the flow-based diffusion pathway improves per-frame refinement quality. A \textbf{Multi-Scale Pose Representation} design captures complementary face, body, and hand dynamics, and a \textbf{Confidence-Aware Causal Attention} mechanism uses keypoint reliability to improve robustness. Our main contributions are summarized as follows:

\begin{itemize}
\item A hybrid autoregressive-diffusion framework for low-latency SLP that combines frame-wise causal generation with flow-based diffusion refinement.
\item A three-expert multi-scale pose representation and fusion design that preserves fine-grained articulator details while explicitly modeling coupled bimanual motion.
\item A confidence-aware causal attention mechanism and a self-forcing training protocol for improving robustness under noisy 2D pose observations.
\end{itemize}

\section{Related Work}

Sign language is a primary means of communication for deaf and hard-of-hearing communities, yet communication barriers remain common because many hearing individuals do not know sign language. Early research therefore focused mainly on Sign Language Recognition (SLR)~\cite{slr1, slr2, slr3, slr4, slr5} and Sign Language Translation (SLT)~\cite{slt1, slt2, slt3}. More recently, increasing attention has been directed toward SLP, which aims to generate sign motion from linguistic input.

\subsection{Sign Language Production}

Initial studies in SLP predominantly relied on rule-based animation techniques to translate textual inputs into synthetic avatar-based sign language animations~\cite{slp1, slp2, slp3}. These methods typically employed predefined templates and handcrafted lookup rules to map sentences or glosses to gestures. While these systems offered interpretable sign animations, they suffered from high data collection costs, limited scalability, and poor generalization to unseen sentences or grammatical structures.

\subsection{Autoregressive models in SLP}

With the advent of deep learning, more flexible and data-driven SLP models were introduced. Text2Sign~\cite{text2sign} introduces a multi-stage pipeline for SLP, which decomposes the overall process into three sequential subtasks: text-to-gloss\footnote{Gloss is a written representation of signs using capitalized words to show the meaning and order of the signs.} translation, gloss-to-pose generation, and pose-to-video synthesis. While effective, such pipeline designs introduced potential efficiency issues and error propagation between stages. To address this,~\cite{ptslp} presented the first end-to-end autoregressive model that directly generates sign pose sequences from glosses. Further improvements followed, including a multi-channel adversarial model~\cite{atmc} and a Mixture Density Transformer~\cite{3dmc} to better model the multi-modal nature of sign gestures.

\subsection{Diffusion models in SLP}

Recently, diffusion models have shown strong potential in SLP due to their capacity for modeling complex output distributions. Methods such as SignDiff~\cite{signdiff}, G2P-DDM~\cite{g2pddm}, and GCDM~\cite{gcdm} leverage denoising diffusion probabilistic models (DDPMs~\cite{ddpm}) to generate sign pose sequences by gradually refining Gaussian noise into meaningful joint coordinates. These approaches typically treat pose generation as a coordinate regression problem, guided by gloss or semantic input. Subsequently, Sign-IDD~\cite{signidd} integrates an attribute-controllable diffusion module that leverages skeletal direction and length attributes to constrain joint associations, thereby enabling more precise and controllable pose generation.

\section{Preliminary}

In this section, we introduce the fundamental concepts and notations used in our framework, including autoregressive modeling, diffusion-based generation, human pose representation, and attention mechanisms.

\subsection{Autoregressive Modeling}

Autoregressive models generate a sequence by modeling each element conditioned on its previous elements. For a sequence of human poses $P = \{P_1, P_2, \ldots, P_T\}$, where $P_t \in \mathbb{R}^{J \times D}$ represents the positions of $J$ joints at time $t$ in $D$-dimensional space, the autoregressive factorization is:

\begin{equation}
    p(P) = \prod_{t=1}^{T} p(P_t \mid P_{<t})
\end{equation}

This formulation allows for temporal modeling but is prone to error accumulation due to its greedy generation process.

\subsection{Diffusion Models}

\paragraph{Score-based diffusion models.}
Diffusion models learn to generate data by reversing a gradual noising process. Traditional diffusion models~\cite{sde, ddim, iddpm} follow the denoising diffusion probabilistic model (DDPM) framework~\cite{ddpm}, where a data sample $x_0$ is progressively perturbed over $T$ steps using a predefined noise schedule $\{\beta_t\}_{t=1}^T$. The forward process is defined as:

\begin{equation}
    q(x_t \mid x_{t-1}) = \mathcal{N}(x_t; \sqrt{1 - \beta_t} x_{t-1}, \beta_t I)
\end{equation}

This results in:

\begin{equation}
    q(x_t \mid x_0) = \mathcal{N}(x_t; \sqrt{\bar{\alpha}_t} x_0, (1 - \bar{\alpha}_t) I)
\end{equation}

where $\alpha_t = 1 - \beta_t$ and $\bar{\alpha}_t = \prod_{s=1}^{t} \alpha_s$.

The denoising model is trained to predict the added noise $\boldsymbol{\epsilon}$:

\begin{equation}
    \mathcal{L}_{\text{Diff}} = \mathbb{E}_{x_0, t, \boldsymbol{\epsilon}} \left[ \left\| \boldsymbol{\epsilon} - \boldsymbol{\epsilon}_\theta(x_t, t) \right\|^2 \right]
\end{equation}

However, traditional diffusion models have slow sampling speeds, which significantly hinder low-latency generation.

\paragraph{Flow-based Diffusion Models.}
In contrast to DDPMs, flow-based diffusion models~\cite{flowmatching1, flowmatching2, rectifiedflow} learn an explicit transformation between noise and data using invertible mappings. These models parameterize the denoising process as an ODE (or deterministic flow) rather than a stochastic process. Specifically, they model a continuous-time trajectory between the data distribution and a base noise distribution using a neural network as the vector field.

The process is defined through the following differential equation:

\begin{equation}
\frac{dx(t)}{dt} = v_\theta(x(t), t), \quad x(0) \sim \mathcal{N}(0, I)
\end{equation}

Here, $v_\theta$ is a neural network trained to match the score or displacement between perturbed data and clean data. This flow transports samples from the base distribution (e.g., Gaussian) at $t=0$ to the data distribution at $t=1$.

A popular training objective is the flow matching loss:

\begin{equation}
\mathcal{L}_{\text{Flow}} = \mathbb{E}_{x_0, t, z} \left[ \left\| v_\theta(x_t, t) - \frac{x_0 - x_t}{t} \right\|_2^2 \right]
\end{equation}

where $x_t = (1 - t) z + t x_0$ is a linear interpolation between a noise sample $z \sim \mathcal{N}(0, I)$ and a data sample $x_0$.

This objective encourages the model to learn a displacement field that transports noise into data along straight paths, offering a deterministic alternative to stochastic diffusion, thereby accelerating training efficiency and sampling speed.

\subsection{Human Pose Representation}

A human pose at time $t$ is represented as a set of $J$ joints:
\begin{equation}
    P_t = \{p_{t,1}, p_{t,2}, \ldots, p_{t,J}\}, \quad p_{t,j} \in \mathbb{R}^{D}
\end{equation}

Each joint is also associated with a confidence score $c_{t,j} \in [0, 1]$, denoted as:

\begin{equation}
    C_t = \{c_{t,1}, c_{t,2}, \ldots, c_{t,J} \}
\end{equation}

These confidence scores are used later in our attention mechanism to modulate the influence of each joint during pose generation (Section 4.4).

\subsection{Attention Mechanism}

We build upon the standard scaled dot-product attention mechanism. Given query $Q \in \mathbb{R}^{n \times d}$, key $K \in \mathbb{R}^{m \times d}$, and value $V \in \mathbb{R}^{m \times d}$ matrices, the attention is computed as:

\begin{equation}
    \text{Attention}(Q, K, V) = \text{softmax}\left( \frac{QK^\top}{\sqrt{d}} \right)V
\end{equation}

To preserve temporal causality in generation, we apply a causal mask to prevent attention to future time steps.

Later, in our proposed method, we extend this formulation to include confidence-awareness, biasing the attention distribution based on joint reliability (Section 4.4).
\begin{figure}[ht]
    \centering
    \includegraphics[width=0.98\linewidth]{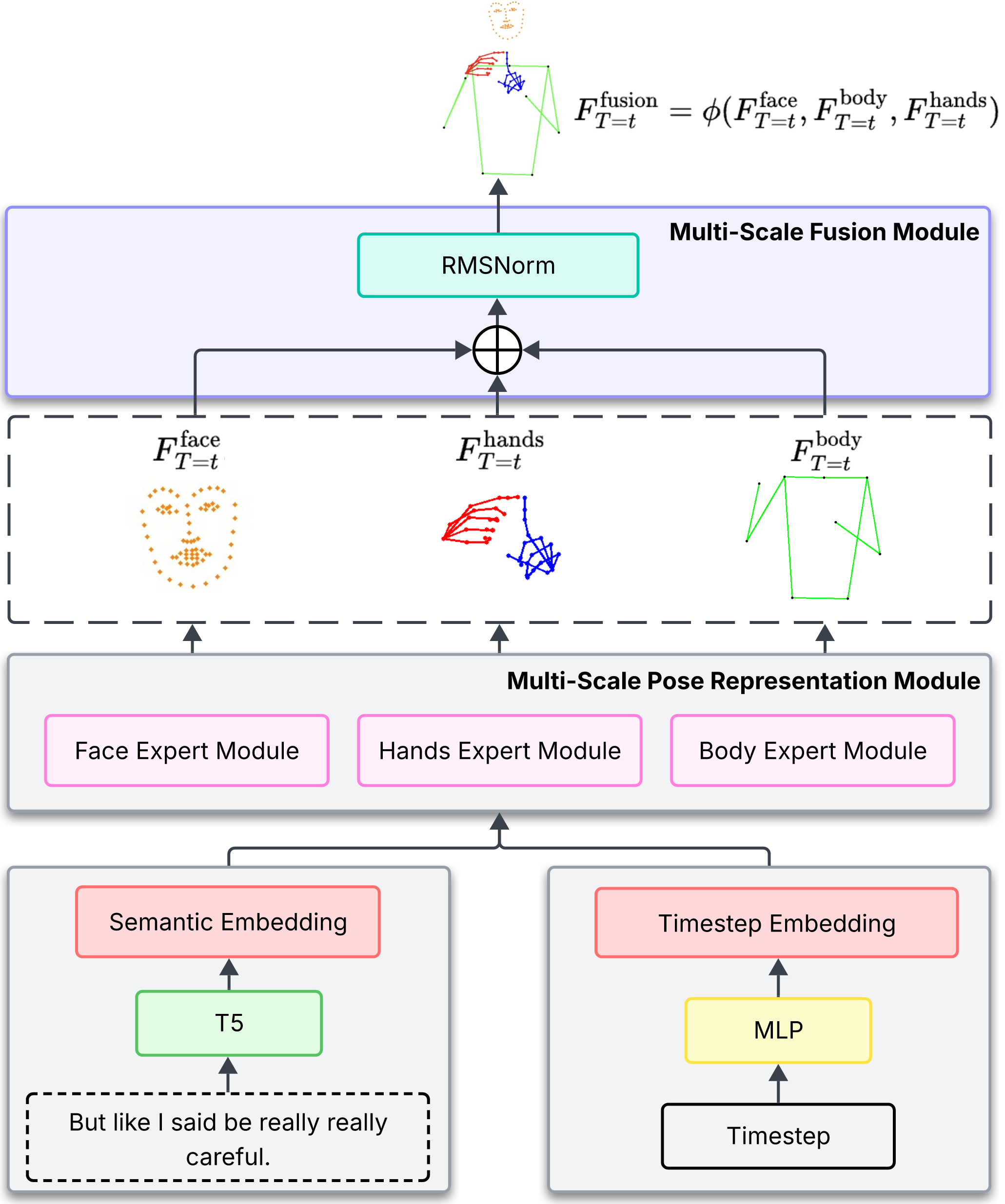}
    \caption{Overview of Hybrid Autoregressive-Diffusion Model.}
    \label{fig:overview}
\end{figure}
\section{Method}





Our framework integrates a Hybrid Autoregressive-Diffusion model, Multi-Scale Pose Representation and Fusion modules, and a Confidence-Aware Causal Attention mechanism for temporally coherent, high-quality sign language pose generation.

\begin{figure}[ht]
    \centering
    \includegraphics[width=0.98\linewidth]{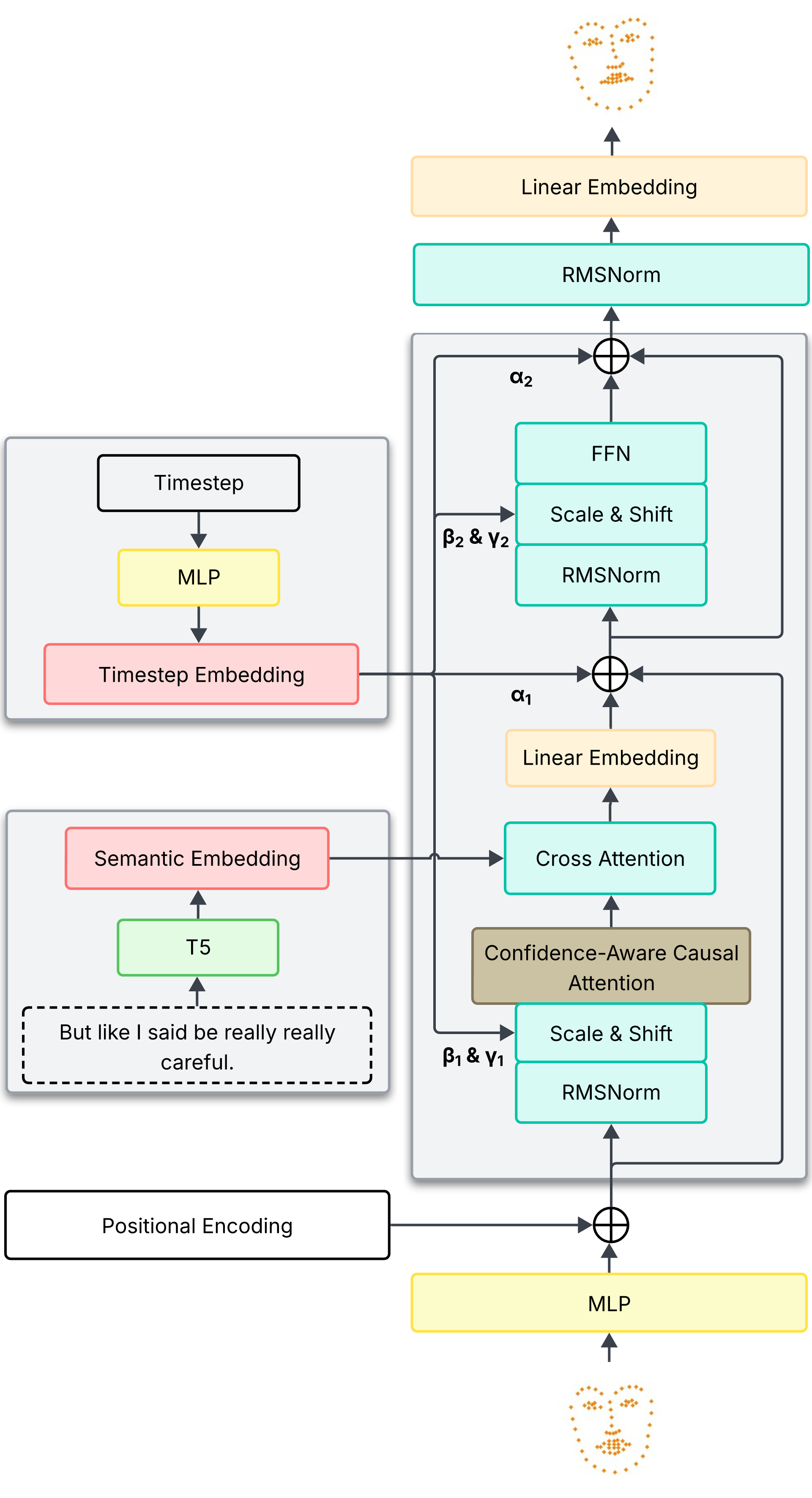}
    \caption{Expert module for facial features in the Multi-Scale Pose Representation Module.}
    \label{fig:expert}
\end{figure}

\subsection{Overview}

As shown in Figure~\ref{fig:overview}, given a natural language sentence, the model generates a temporally consistent pose sequence frame by frame. The Multi-Scale Pose Representation module first produces face, body, and hand articulators via dedicated Expert modules, which are fused into a full pose frame. The fused pose is then decomposed and used as the autoregressive condition for the next frame. Joint confidence scores guide the process to improve accuracy and realism.

\subsection{Hybrid Autoregressive-Diffusion Model}

To the best of our knowledge, this is the first SLP model that combines autoregressive and diffusion paradigms for SLP, aiming to harness the advantages of both.

Specifically, we introduce causal attention into the denoiser of the diffusion model by applying an attention mask $M \in \mathbb{R}^{n \times n}$ to its self-attention mechanism: 

\begin{equation}
M_{ij} =
\begin{cases}
0 & \text{if } j \leq i \\
-\infty & \text{if } j > i
\end{cases}
\end{equation}

so that each position $i$ can only attend to positions $j \leq i$ (i.e., current and previous tokens).

\begin{figure*}[ht]
    \centering
    \includegraphics[width=0.98\linewidth]{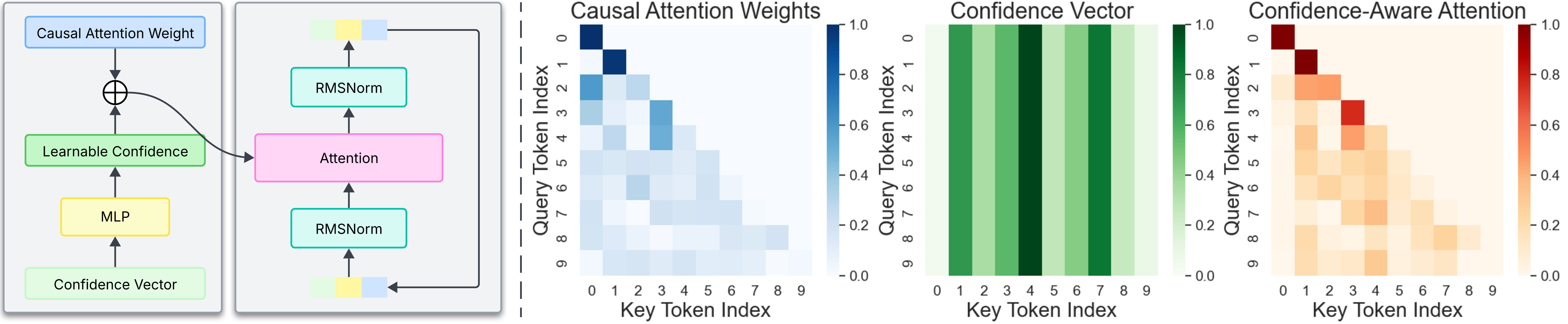}
    \caption{Confidence-Aware Causal Attention Mechanism.}
    \label{fig:confidence-aware}
\end{figure*}
To address the distribution mismatch between training and inference commonly observed in autoregressive models, we draw on the training strategy proposed in Self-Forcing~\cite{selfforing}. Instead of relying on ground truth during training, our model always conditions the next step on its own previously generated articulators. Specifically, at each time step $t$, the model produces

\begin{equation}
\begin{aligned}
\tilde{A}_t &= f_{\text{AR}}(\tilde{A}_{<t}), \\
\tilde{A}_{<t} &= \{\tilde{A}_1, \dots, \tilde{A}_{t-1}\}, \\
\tilde{A}_t &= \{\tilde{F}_t, \tilde{H}_t, \tilde{B}_t\}.
\end{aligned}
\end{equation}

where $\tilde{F}_t$, $\tilde{H}_t$, and $\tilde{B}_t$ denote the generated face, hands, and body components at time step $t$, respectively. These components are fused into a full-frame pose and then decomposed again to serve as the condition for the next step. Using $\tilde{A}_t$ for the articulator tuple avoids overloading the confidence notation $C_t$ introduced in Section~3.3.

\paragraph{Self-forcing protocol.} For each sentence, the first frame is generated from the text condition alone. For $t>1$, the three experts generate the current face, hand, and body components in parallel, conditioned on the decomposed prediction from time $t-1$. During training, we do not replace this autoregressive condition with the ground truth. As a result, training and inference share the same conditioning path, which substantially reduces exposure bias. The holistic Soft-DTW loss~\cite{softdtw} is applied on the generated sequence to provide sequence-level supervision and further mitigate error accumulation.

To improve training efficiency and sampling speed, we adopt a flow-based diffusion model~\cite{flowmatching1, flowmatching2} instead of the traditional DDPM-based approach~\cite{ddpm, ddim}. Flow matching enables learning a continuous transformation path that maps the initial distribution directly to the target distribution, allowing sampling to be completed in fewer steps and enhancing both training and sampling efficiency.

\subsection{Multi-Scale Pose Representation Module}

Sign language involves coordinated motion across face, body, and hands. To capture this, we design a \textbf{Multi-Scale Pose Representation} module (Figure~\ref{fig:expert}) that partitions keypoints into three anatomical groups and processes each with a dedicated expert module.

Given a sequence of $T$ frames with $N$ keypoints per frame, each keypoint $p_t^{(i)} = [x_t^{(i)}, y_t^{(i)}, c_t^{(i)}]$ is embedded as:

\begin{equation}
s_t^{(i)} = \mathrm{MLP}(p_t^{(i)}) + PE(t),
\end{equation}

We divide the keypoints into groups $\mathcal{G} = {G_{\text{face}}, G_{\text{body}}, G_{\text{hands}}}$ and extract group-specific features:

\begin{equation}
S_t^{(k)} = { s_t^{(i)} \mid i \in G_k }, \quad k = 1,2,3
\end{equation}

Each group sequence is processed by a scale-specific expert $\mathcal{E}^{(k)}$ to capture intra-scale dependencies:

\begin{equation}
H^{(k)} = \mathcal{E}^{(k)}([S_1^{(k)}, \dots, S_T^{(k)}])
\end{equation}

We average-pool over joints and apply attention-based fusion across scales:

\begin{equation}
\begin{aligned}
H_t &= \sum_{k=1}^{3} \alpha_t^{(k)} \, H_t^{(k)}, \\
\alpha_t^{(k)} &= 
\frac{
    \exp\Big(w^\top \tanh\big(w_f H_t^{(k)}\big)\Big)
}{
    \sum_{j=1}^{3} \exp\Big(w^\top \tanh\big(w_f H_t^{(j)}\big)\Big)
}.
\end{aligned}
\end{equation}

The fused representation $H_t$ serves as input for subsequent generation modules.

\subsection{Confidence-Aware Causal Attention Mechanism}

To improve robustness in autoregressive pose generation, we propose a \textbf{Confidence-Aware Causal Attention} module (Figure~\ref{fig:confidence-aware}), which integrates keypoint confidence scores into the attention computation to downweight unreliable inputs.

Given a sentence $S$, the model generates a pose sequence $x = {x^{(1)}, \dots, x^{(T)}}$, where each $x^{(t)} \in \mathbb{R}^{J \times d}$ represents $J$ keypoints. Each keypoint has an associated confidence score $c_i^{(t)} \in [0,1]$.
\begin{table*}[ht]
\centering
\resizebox{0.98\textwidth}{!}{
\begin{tabular}{l|cccccc|cccccc}
\toprule
\multirow{2}{*}{Methods} & \multicolumn{6}{c|}{\textbf{DEV}} & \multicolumn{6}{c}{\textbf{TEST}} \\
 & B1$\uparrow$ & B4$\uparrow$ & ROUGE$\uparrow$ & WER$\downarrow$ & DTW$\downarrow$ & FID$\downarrow$ & B1$\uparrow$ & B4$\uparrow$ & ROUGE$\uparrow$ & WER$\downarrow$ & DTW$\downarrow$ &FID$\downarrow$ \\
\midrule
PT~\cite{ptslp} & 12.51 & 3.88 & 11.87 & 96.85 & - & - & 13.35 & 4.31 & 13.17 & 96.50 & - & - \\
G2P-DDM~\cite{g2pddm} & - & - & - & - & - & - & 16.11 & 7.50 & - & 77.26 & - & - \\
GCDM~\cite{gcdm} & 22.88 & 7.64 & 23.35 & 82.81 & 11.18 & 39.87 & 22.03 & 7.91 & 23.20 & 81.94 & 11.10 & 49.22 \\
GEN-OBT~\cite{gen-obt} & 24.92 & 8.68 & 25.21 & 82.36 & - & - & 23.08 & 8.01 & 23.49 & 81.78 & - & - \\
Sign-IDD~\cite{signidd} & 25.40 & 8.93 & 27.60 & 77.72 & 5.09 & 39.11 & 24.80 & 9.08 & 26.58 & 76.66 & 6.20 & 47.19 \\
\textbf{HybridSign (Ours)} & \textbf{26.98} & \textbf{9.26} & \textbf{28.07} & \textbf{75.81} & \textbf{4.07} & \textbf{38.28} & \textbf{25.77} & \textbf{10.03} & \textbf{27.97} & \textbf{75.02} & \textbf{4.96} & \textbf{45.50} \\
\midrule
Ground Truth & 29.77 & 12.13 & 29.60 & 74.17 & 0.00 & 0.00 & 29.76 & 11.93 & 28.98 & 71.94 & 0.00 & 0.00 \\
\bottomrule
\end{tabular}
}
\caption{Performance comparison on PHOENIX14T. Higher B1/B4/ROUGE indicates better back-translation quality, while lower WER/DTW/FID indicates better motion fidelity and temporal alignment.}
\label{table:performance-on-phoenix14t}
\end{table*}

\begin{table*}[ht]
\centering
\resizebox{0.98\textwidth}{!}{
\begin{tabular}{l|cccccc|cccccc}
\toprule
\multirow{2}{*}{Methods} & \multicolumn{6}{c|}{\textbf{DEV}} & \multicolumn{6}{c}{\textbf{TEST}} \\
 & B1$\uparrow$ & B4$\uparrow$ & ROUGE$\uparrow$ & WER$\downarrow$ & DTW$\downarrow$ & FID$\downarrow$ & B1$\uparrow$ & B4$\uparrow$ & ROUGE$\uparrow$ & WER$\downarrow$ & DTW$\downarrow$ &FID$\downarrow$ \\
\midrule
PT~\cite{ptslp} & 14.34 & 4.07 & 8.12 & 96.91 & 10.53 & 55.02 & 14.05 & 4.12 & 8.42 & 96.47 & 10.18 & 54.57 \\
G2P-DDM~\cite{g2pddm} & 19.82 & 5.37 & 12.47 & 90.05 & 8.25 & 50.48 & 19.48 & 5.12 & 12.21 & 89.58 & 7.97 & 49.83 \\
GCDM~\cite{gcdm} & 26.43 & 5.84 & 15.53 & 91.92 & 6.32 & 46.19 & 25.91 & 5.57 & 15.21 & 91.43 & 6.13 & 45.71 \\
GEN-OBT~\cite{gen-obt} & 28.63 & 6.14 & 16.23 & 91.07 & 7.08 & 48.03 & 27.82 & 5.92 & 15.88 & 90.63 & 6.87 & 47.28 \\
Sign-IDD~\cite{signidd} & 29.12 & 6.27 & 17.01 & 89.95 & 4.76 & 35.03 & 28.90 & 6.06 & 16.21 & 89.98 & 4.86 & 39.02 \\
\textbf{HybridSign (Ours)} & \textbf{30.71} & \textbf{6.92} & \textbf{18.96} & \textbf{87.11} & \textbf{3.76} & \textbf{34.19} & \textbf{30.12} & \textbf{6.48} & \textbf{18.02} & \textbf{88.30} & \textbf{3.89} & \textbf{37.10} \\
\midrule
Ground Truth & 35.20 & 8.89 & 22.45 & 83.79 & 0.00 & 0.00 & 34.01 & 8.03 & 21.87 & 81.94 & 0.00 & 0.00 \\
\bottomrule
\end{tabular}
}
\caption{Performance comparison on How2Sign. Higher B1/B4/ROUGE indicates better back-translation quality, while lower WER/DTW/FID indicates better motion fidelity and temporal alignment.}
\label{table:performance-on-how2sign}
\end{table*}

\paragraph{Confidence-Weighted Attention.}

In causal attention, timestep $t$ only attends to ${1, \dots, t}$:

\begin{equation}
\alpha_{t,s} = \frac{\exp\left( \frac{Q_t K_s^\top}{\sqrt{d_k}} \right)}{\sum_{j=1}^{t} \exp\left( \frac{Q_t K_j^\top}{\sqrt{d_k}} \right)}
\end{equation}

We introduce a confidence bias to this formulation:

\begin{equation}
\alpha_{t,s} = \frac{
\exp\left( \frac{Q_t K_s^\top}{\sqrt{d_k}} + \beta \cdot \bar{c}^{(s)} \right)
}{
\sum_{j=1}^{t} \exp\left( \frac{Q_t K_j^\top}{\sqrt{d_k}} + \beta \cdot \bar{c}^{(j)} \right)
}
\end{equation}

where $\bar{c}^{(s)} = \frac{1}{J} \sum_{i=1}^{J} c_i^{(s)}$ is the average keypoint confidence, and $\beta$ is a learnable scalar controlling the strength of the bias.

This allows the model to attend more to reliable frames during decoding.

\subsection{Training Objectives}
Our model is optimized using a composite loss that combines three complementary objectives: joint accuracy, kinematic consistency, and temporal alignment.

\paragraph{Joint loss}
To encourage accurate joint localization, we minimize the mean absolute error between predictions and ground truth:
\begin{equation}
\mathcal{L}_{\mathrm{joint}} \;=\;
\frac{1}{J}\sum_{j=1}^{J} \| p_j - \hat{p}_j \|_1,
\end{equation}
where \(p_j\in\mathbb{R}^d\) and \(\hat{p}_j\in\mathbb{R}^d\) denote the ground-truth and predicted position of joint \(j\), and \(\|\cdot\|_1\) is the \(\ell_1\) norm.

\paragraph{Bone loss}
To preserve bone orientations and kinematic consistency, we penalize orientation deviations:
\begin{equation}
\mathcal{L}_{\mathrm{bone}} \;=\;
\frac{1}{B}\sum_{b=1}^{B} \| q_b - \hat{q}_b \|_2^2,
\end{equation}
where \(q_b\) and \(\hat{q}_b\) are the ground-truth and predicted orientation (e.g. unit quaternions or axis-angle vectors) of bone \(b\), and \(\|\cdot\|_2\) denotes the Euclidean norm. If orientations are represented as quaternions, consider using a geodesic/angular metric as an alternative.

\paragraph{Soft-DTW loss}
To align predicted and ground-truth pose sequences temporally, we employ differentiable Dynamic Time Warping (Soft-DTW)~\cite{softdtw}:
\begin{equation}
\mathcal{L}_{\mathrm{soft\text{-}dtw}} \;=\;
\min\nolimits_{A}^{\gamma} \sum_{(i,j)\in A} \| x_i - y_j \|_2^2,
\end{equation}
where \(x_i\) and \(y_j\) are per-frame pose descriptors, \(A\) is an alignment path, and \(\gamma>0\) controls the smoothness of the soft minimum:
\begin{equation}
\min\nolimits^{\gamma}\{a_1,\dots,a_k\}
= -\gamma \log\!\Bigg(\sum_{i=1}^k e^{-a_i/\gamma}\Bigg).
\end{equation}

\paragraph{Adaptive total loss}
We balance the three terms with dynamic weights computed from the inverse exponential moving average (EMA) of each loss~\cite{dwa}. Let \(\bar{\mathcal{L}}_i^{(t)}\) be the EMA of loss \(i\) at iteration \(t\), and \(\epsilon>0\) a small constant to avoid division by zero. The weight for loss \(i\) is
\begin{equation}
\lambda_i^{(t)} \;=\;
\frac{\big(\bar{\mathcal{L}}_i^{(t)} + \epsilon\big)^{-1}}
{\sum_{j=1}^{N} \big(\bar{\mathcal{L}}_j^{(t)} + \epsilon\big)^{-1}},
\qquad i=1,\dots,N,
\end{equation}
with \(N=3\) in our case. The final objective at iteration \(t\) is
\begin{equation}
\mathcal{L}_{\mathrm{total}}^{(t)} \;=\;
\lambda_1^{(t)}\,\mathcal{L}_{\mathrm{joint}}^{(t)} \;+\;
\lambda_2^{(t)}\,\mathcal{L}_{\mathrm{bone}}^{(t)} \;+\;
\lambda_3^{(t)}\,\mathcal{L}_{\mathrm{soft\text{-}dtw}}^{(t)}.
\end{equation}

This adaptive weighting scheme emphasizes losses that are relatively smaller (via inverse EMA), promoting stable and balanced training across spatial accuracy, kinematic consistency, and temporal alignment.

\section{Experiment}
\begin{figure*}[htbp]
    \centering
    \includegraphics[width=0.98\linewidth]{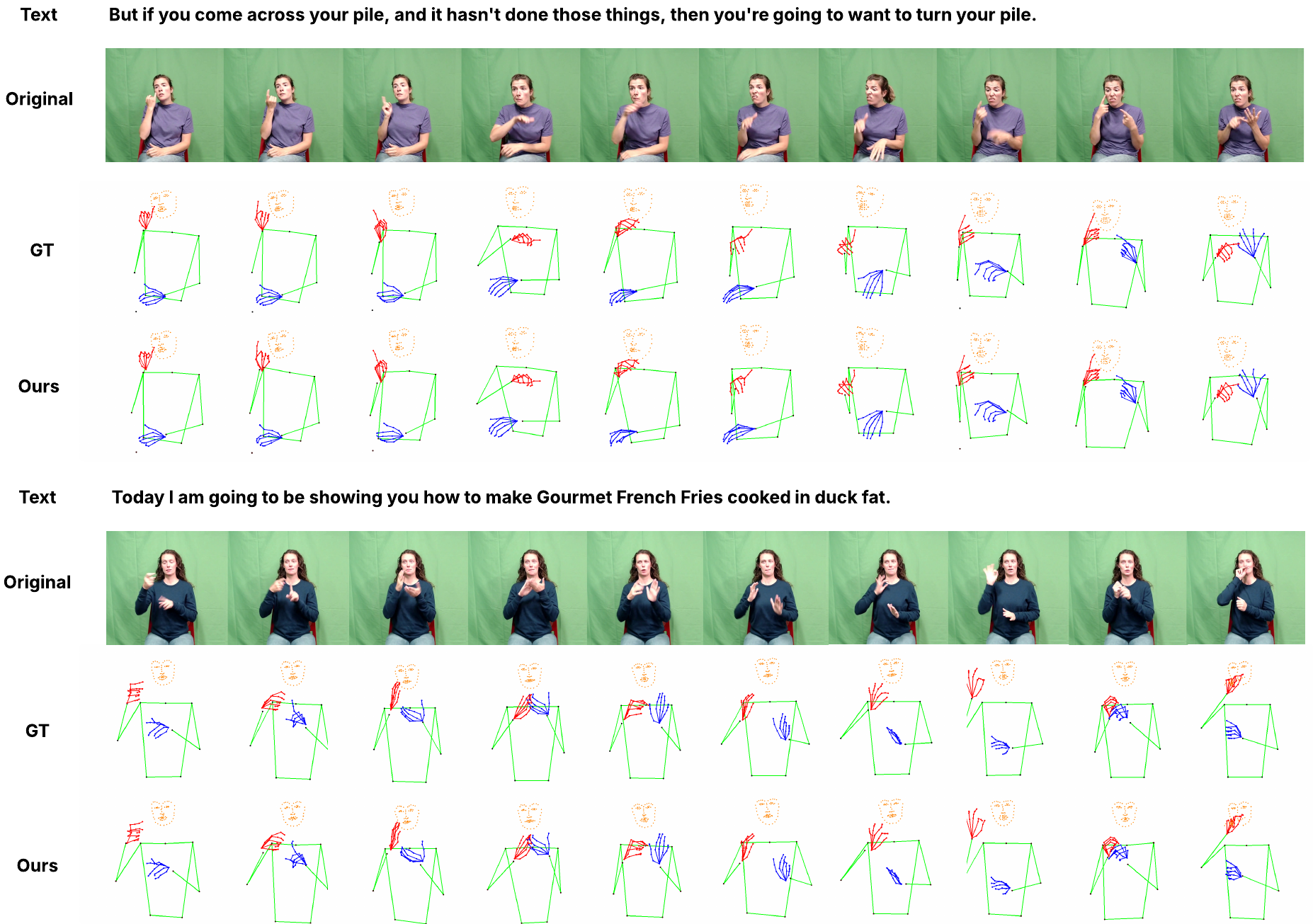}
    \caption{Visualization examples of generated poses on How2Sign. We compare HybridSign with the ground-truth poses and the original video frames for clear evaluation.}
    \label{fig:comparison}
\end{figure*}

\subsection{Experimental Setup}

\paragraph{Datasets.} 
We evaluate the proposed method on two benchmarks: PHOENIX14T~\cite{phoenix} and How2Sign~\cite{how2sign}. PHOENIX14T is a German Sign Language corpus that consists of 8,257 sentence-level samples and 2,887 unique German words. How2Sign is a large-scale American Sign Language corpus designed for sign language understanding and generation tasks, containing more than 80 hours of multimodal data.

\paragraph{Evaluation Metrics.} 
Following the existing works~\cite{g2pddm, gcdm, signidd, signd2c}, we used a pre-trained SLT model~\cite{slt} for back-translation, converting generated sign language pose sequences back to text, and then evaluated the results with BLEU~\cite{bleu}, ROUGE~\cite{rouge}, WER, DTW, and FID.

\paragraph{Training and inference protocol.}
Unless otherwise stated, all models generate sequences of 60 frames. The three articulator experts are executed in parallel within each time step, while the autoregressive dependency is imposed only across time. We report \emph{time-to-first-frame} as latency, i.e., the wall-clock time required to output the first pose frame of a 60-frame sequence, because this quantity best reflects interactive usability. Throughout the paper, our use of the term \emph{real-time} therefore refers to this low-latency, continuous generation setting relative to prior diffusion-based SLP systems rather than to instantaneous end-to-end video synthesis. During both training and inference, the previous pose fed to the next step is always the model prediction rather than the ground truth, which keeps the conditioning distribution consistent with the self-forcing design described in Section~4.2. Unless otherwise stated, the ablation results in Tables~\ref{table:ablation-mode}--\ref{table:ablation-expert} are reported on the How2Sign test split.

\subsection{Comparisons with Baseline Models}

We conducted both qualitative and quantitative analyses to evaluate the model's performance.

\begin{table}[ht]
\centering
\resizebox{0.48\textwidth}{!}{
\begin{tabular}{l|cc}
\toprule
Method & Latency (s)$\downarrow$ & Throughput (FPS)$\uparrow$ \\
\midrule
GCDM~\cite{gcdm} & 52.18 & 1.15 \\
Sign-IDD~\cite{signidd} & 40.31 & 1.49 \\
G2P-DDM~\cite{g2pddm} & 25.78 & 2.33 \\
\textbf{HybridSign} & \textbf{5.90} & \textbf{10.17}  \\
\bottomrule
\end{tabular}
}
\caption{Latency and throughput comparison under the 60-frame evaluation protocol. Latency measures time-to-first-frame rather than the total sequence generation time.}
\label{table:latency}
\end{table}

\paragraph{Quantitative Analysis.}
Tables~\ref{table:performance-on-phoenix14t} and~\ref{table:performance-on-how2sign} present a comparison of HybridSign and the baseline models on PHOENIX14T and How2Sign, respectively. Across both datasets, HybridSign delivers the strongest overall quality--efficiency trade-off and achieves the best results among the compared methods on the reported metrics.

Table~\ref{table:latency} compares latency and throughput with the baseline models. Here, latency refers to the time required to generate the first frame of the 60-frame sequence, rather than the total time to produce the entire video. This reflects the model's initial response speed, which is especially important for interactive deployment where a system should start responding before the entire sequence is synthesized. HybridSign achieves substantially lower latency and higher throughput than the compared diffusion-based baselines, indicating a strong quality--efficiency trade-off for low-latency sign language production.

Performance differences across the two datasets are relatively small, which is partly attributable to our use of a unified 2D pose representation across both benchmarks. This choice improves cross-dataset consistency and computational efficiency, but it also removes depth cues that are important for subtle hand-face interactions and for disambiguating overlapping articulators. Our Multi-Scale Pose Representation and Confidence-Aware Causal Attention mitigate this issue by emphasizing reliable local structure, yet they do not fully recover missing 3D information. Notably, the DTW score is reduced by approximately 20\% thanks to the Soft-DTW loss~\cite{softdtw}, which improves temporal alignment and stabilizes long-horizon autoregressive generation.

\paragraph{Qualitative Analysis.}
Figure~\ref{fig:comparison} illustrates representative generated poses alongside the corresponding ground truth, and Figure~\ref{fig:comparison-diverse} further expands the qualitative evaluation with more diverse examples. Across both figures, HybridSign preserves the global motion trend, hand trajectories, and major body posture changes. The most visible deviations occur in subtle bone lengths, fine wrist angles, and local configurations when multiple articulators become spatially close. Even in these cases, the generated poses remain semantically interpretable, which is consistent with the strong back-translation scores in Tables~\ref{table:performance-on-phoenix14t} and~\ref{table:performance-on-how2sign}.

\begin{table}[ht]
\centering
\resizebox{0.48\textwidth}{!}{
\begin{tabular}{l|ccc|cc}
\toprule
Modes & B1$\uparrow$ & B4$\uparrow$ & DTW$\downarrow$ & Latency (s)$\downarrow$ & Throughput (FPS)$\uparrow$ \\
\midrule
Diffusion Mode & \textbf{30.25} & \textbf{6.55} & 8.06 & 32.89 & 1.83 \\
Autoregressive Mode & 26.15 & 5.40 & 4.49 & \textbf{5.53} & \textbf{10.85} \\
\textbf{Hybrid Mode} & \underline{30.12} & \underline{6.48} & \textbf{3.89} & \underline{5.90} & \underline{10.17} \\
\bottomrule
\end{tabular}
}
\caption{Ablation study results of generation modes on the How2Sign test split.}
\label{table:ablation-mode}
\end{table}

\begin{table}[ht]
\centering
\resizebox{0.48\textwidth}{!}{
\begin{tabular}{l|ccc|cc}
\toprule
Modules & B1$\uparrow$ & B4$\uparrow$ & DTW$\downarrow$ & Latency (s)$\downarrow$ & Throughput (FPS)$\uparrow$ \\
\midrule
RNN & 23.47 & 5.02 & 6.89 & \textbf{4.72} & 9.71 \\
CA$^{\dagger}$ & 25.08 & 5.83 & 5.50 & 5.48 & \textbf{10.95} \\
\textbf{CACA$^{\ddagger}$} & \textbf{30.12} & \textbf{6.48} & \textbf{3.89} & \underline{5.90} & \underline{10.17}\\
\bottomrule
\end{tabular}
}
\caption{Ablation study results of autoregressive backbones on the How2Sign test split. CA$^{\dagger}$ denotes Causal Attention, and CACA$^{\ddagger}$ denotes Confidence-Aware Causal Attention.}
\label{table:ablation-module}
\end{table}

\begin{table}[ht]
\centering
\resizebox{0.48\textwidth}{!}{
\begin{tabular}{l|ccc|cc}
\toprule
Experts & B1$\uparrow$ & B4$\uparrow$ & DTW$\downarrow$ & Latency (s)$\downarrow$ & Throughput (FPS)$\uparrow$ \\
\midrule
1 (whole pose) & 22.33 & 5.14 & 7.02 & 7.69 & 7.80 \\
4 (face + body + lh + rh) & 29.17 & 6.03 & 5.72 & 7.73 & 7.76 \\
\textbf{3 (face + body + hands)} & \textbf{30.12} & \textbf{6.48} & \textbf{3.89} & \textbf{5.90} & \textbf{10.17}\\
\bottomrule
\end{tabular}
}
\caption{Ablation study results of different expert decompositions on the How2Sign test split.}
\label{table:ablation-expert}
\end{table}

\subsection{Ablation Study}

We conducted three ablation studies on the generation of 60-frame sign language pose sequences to systematically evaluate our method. Unless otherwise stated, the ablations are reported on the How2Sign test split. Specifically, we compared the proposed hybrid approach with a pure diffusion model and a pure autoregressive model, and further investigated the impact of different attention mechanisms and varying numbers of expert modules.

Table ~\ref{table:ablation-mode} compares the performance and efficiency of a pure diffusion model and a pure autoregressive model. Specifically, the diffusion model adopts the DiT architecture ~\cite{dit}, while the autoregressive model uses a Transformer decoder with teacher forcing ~\cite{attention}. Experimental results show that, thanks to its multi-step denoising process, the diffusion model achieves very high generation quality and reaches the best scores on both the B1 and B4 metrics. However, its temporal consistency and generation efficiency lag significantly behind the other models, making it unsuitable for low-latency scenarios. In contrast, the autoregressive model generates frame-by-frame, resulting in extremely low first-frame latency, but suffers from a noticeable quality drop due to the mismatch between training and inference. Our proposed approach combines the high-quality generation of diffusion models with the frame-by-frame generation of autoregressive models, achieving both strong performance and efficiency.

Table~\ref{table:ablation-module} presents the performance differences among various autoregressive implementations, including a recurrent neural network (RNN)~\cite{rnn}, a standard causal attention mechanism~\cite{attention}, and a confidence-aware causal attention mechanism. The results indicate that the RNN-based model achieves very low generation quality, but due to its computational simplicity compared to attention mechanisms, it exhibits excellent first-frame latency. However, its strict sequential dependency limits overall efficiency, resulting in lower throughput than attention-based methods. In contrast, the confidence-aware causal attention mechanism incorporates confidence scores, assigning different attention weights based on the reliability of keypoints, which leads to the best generation quality while maintaining strong applicability in low-latency scenarios.

Table~\ref{table:ablation-expert} presents the results of varying the number of expert modules. The experiments include using a single expert module (treating the entire pose as a whole), using three expert modules to handle the face, body, and hands separately, and using four expert modules to handle the face, body, left hand, and right hand individually. The single-expert setting cannot capture local articulator-specific details, leading to a significant drop in generation quality. More importantly, the comparison between three and four experts reveals that finer decomposition is not always better. In sign language, the two hands form a strongly coupled subsystem: many signs rely on symmetry, anti-symmetry, relative hand distance, and precise cross-hand timing. A unified hands expert can model these bimanual dependencies directly within one latent space. By contrast, splitting the hands into two experts forces the fusion stage to reconstruct cross-hand relations only after separate generation, which weakens explicit modeling of relative geometry and increases the chance of temporal misalignment. This explains why the four-expert variant underperforms the three-expert design in both quality and efficiency. Therefore, three experts provide the best trade-off, balancing local specialization with coherent bimanual coordination.

\subsection{Discussion on 2D Pose Representation}
Our model adopts 2D poses as a pragmatic representation to keep preprocessing consistent across PHOENIX14T and How2Sign and to avoid introducing dataset-specific 3D supervision requirements. This design is effective for large-scale benchmarking, but it also exposes a clear limitation: depth ambiguity is collapsed in the 2D projection. In practice, the most challenging cases arise when the hands approach the face, when one hand occludes the other, or when similar 2D projections correspond to different 3D articulations. The qualitative examples in Figures~\ref{fig:comparison} and~\ref{fig:comparison-diverse} suggest that HybridSign is robust for dominant planar motion, but still exhibits larger local deviations in precisely these ambiguous cases. We therefore view 3D-aware sign production as an important next step: our hybrid autoregressive-diffusion framework is agnostic to the pose dimensionality and can naturally benefit from richer 3D or multi-view pose annotations when such data become available.

\section{Conclusions}

We introduce a hybrid autoregressive-diffusion framework for low-latency SLP that combines temporal modeling with high-quality refinement. Multi-scale pose representation and confidence-aware causal attention improve both accuracy and robustness. Experiments on PHOENIX14T and How2Sign validate the effectiveness of the approach in both generation quality and efficiency under a low-latency evaluation setting.

\section{Limitations}

While our hybrid autoregressive-diffusion framework achieves state-of-the-art performance for low-latency SLP, several limitations remain. First, the approach depends on annotated sign language datasets, which are still limited in size and diversity, potentially restricting generalization to less-represented sign languages or signing styles. Second, although the multi-scale representation captures hand and facial details, subtle finger articulations and non-manual signals (e.g., eye gaze and mouth gestures) are not yet fully modeled. Finally, while the current inference speed is promising for interactive applications, further optimization may be required for deployment on resource-constrained devices.

\section*{Acknowledgments}
Manoharan and Ye were supported in part by the Smart Ideas
(UOA2493, Developing a Reo Turi Interpreter for Ngati
Turi/Sign Language Interpreter Using Weighted Multimodal
Network for Mahuta ki Tai) funded by the MBIE, New Zealand.


\bibliography{custom}

@InProceedings{phoenix,
author = {Camgoz, Necati Cihan and Hadfield, Simon and Koller, Oscar and Ney, Hermann and Bowden, Richard},
title = {Neural Sign Language Translation},
booktitle = {Proceedings of the IEEE Conference on Computer Vision and Pattern Recognition (CVPR)},
month = {June},
year = {2018}
}

@inproceedings{how2sign,
  title     = {{How2Sign: A Large-scale Multimodal Dataset for Continuous American Sign Language}},
  author    = {Duarte, Amanda and Palaskar, Shruti and Ventura, Lucas and Ghadiyaram, Deepti and DeHaan, Kenneth and
               Metze, Florian and Torres, Jordi and Giro-i-Nieto, Xavier},
  booktitle = {Conference on Computer Vision and Pattern Recognition (CVPR)},
  year      = {2021}
}

@article{openpose,
  author  = {Z. {Cao} and G. {Hidalgo Martinez} and T. {Simon} and S. {Wei} and Y. A. {Sheikh}},
  journal = {IEEE Transactions on Pattern Analysis and Machine Intelligence},
  title   = {OpenPose: Realtime Multi-Person 2D Pose Estimation using Part Affinity Fields},
  year    = {2019}
}

@misc{rnn,
      title={Sequence Transduction with Recurrent Neural Networks}, 
      author={Alex Graves},
      year={2012},
      eprint={1211.3711},
      archivePrefix={arXiv},
      primaryClass={cs.NE},
      url={https://arxiv.org/abs/1211.3711} 
}

@misc{attention,
      title={Attention Is All You Need}, 
      author={Ashish Vaswani and Noam Shazeer and Niki Parmar and Jakob Uszkoreit and Llion Jones and Aidan N. Gomez and Lukasz Kaiser and Illia Polosukhin},
      year={2017},
      eprint={1706.03762},
      archivePrefix={arXiv},
      primaryClass={cs.CL},
      url={https://arxiv.org/abs/1706.03762}, 
}

@misc{dwa,
      title={End-to-End Multi-Task Learning with Attention}, 
      author={Shikun Liu and Edward Johns and Andrew J. Davison},
      year={2019},
      eprint={1803.10704},
      archivePrefix={arXiv},
      primaryClass={cs.CV},
      url={https://arxiv.org/abs/1803.10704}, 
}

@inproceedings{bleu,
  title     = {{B}leu: a Method for Automatic Evaluation of Machine Translation},
  author    = {Papineni, Kishore  and
               Roukos, Salim  and
               Ward, Todd  and
               Zhu, Wei-Jing},
  editor    = {Isabelle, Pierre  and
               Charniak, Eugene  and
               Lin, Dekang},
  booktitle = {Proceedings of the 40th Annual Meeting of the Association for Computational Linguistics},
  month     = jul,
  year      = {2002},
  address   = {Philadelphia, Pennsylvania, USA},
  publisher = {Association for Computational Linguistics},
  url       = {https://aclanthology.org/P02-1040/},
  doi       = {10.3115/1073083.1073135},
  pages     = {311--318}
}

@misc{sde,
      title={Score-Based Generative Modeling through Stochastic Differential Equations}, 
      author={Yang Song and Jascha Sohl-Dickstein and Diederik P. Kingma and Abhishek Kumar and Stefano Ermon and Ben Poole},
      year={2021},
      eprint={2011.13456},
      archivePrefix={arXiv},
      primaryClass={cs.LG},
      url={https://arxiv.org/abs/2011.13456}, 
}

@misc{ddim,
      title={Denoising Diffusion Implicit Models}, 
      author={Jiaming Song and Chenlin Meng and Stefano Ermon},
      year={2020},
      eprint={2010.02502},
      archivePrefix={arXiv},
      primaryClass={cs.LG},
      url={https://arxiv.org/abs/2010.02502}, 
}

@misc{iddpm,
      title={Improved Denoising Diffusion Probabilistic Models}, 
      author={Alex Nichol and Prafulla Dhariwal},
      year={2021},
      eprint={2102.09672},
      archivePrefix={arXiv},
      primaryClass={cs.LG},
      url={https://arxiv.org/abs/2102.09672}, 
}

@misc{rectifiedflow,
      title={Flow Straight and Fast: Learning to Generate and Transfer Data with Rectified Flow}, 
      author={Xingchao Liu and Chengyue Gong and Qiang Liu},
      year={2022},
      eprint={2209.03003},
      archivePrefix={arXiv},
      primaryClass={cs.LG},
      url={https://arxiv.org/abs/2209.03003}, 
}

@misc{softdtw,
      title={Soft-DTW: a Differentiable Loss Function for Time-Series}, 
      author={Marco Cuturi and Mathieu Blondel},
      year={2018},
      eprint={1703.01541},
      archivePrefix={arXiv},
      primaryClass={stat.ML},
      url={https://arxiv.org/abs/1703.01541}, 
}

@inproceedings{rouge,
  title     = {{ROUGE}: A Package for Automatic Evaluation of Summaries},
  author    = {Lin, Chin-Yew},
  booktitle = {Text Summarization Branches Out},
  month     = jul,
  year      = {2004},
  address   = {Barcelona, Spain},
  publisher = {Association for Computational Linguistics},
  url       = {https://aclanthology.org/W04-1013/},
  pages     = {74--81}
}

@misc{slt,
  title         = {Sign Language Transformers: Joint End-to-end Sign Language Recognition and Translation},
  author        = {Necati Cihan Camgoz and Oscar Koller and Simon Hadfield and Richard Bowden},
  year          = {2020},
  eprint        = {2003.13830},
  archiveprefix = {arXiv},
  primaryclass  = {cs.CV},
  url           = {https://arxiv.org/abs/2003.13830}
}

@misc{ptslp,
  title         = {Progressive Transformers for End-to-End Sign Language Production},
  author        = {Ben Saunders and Necati Cihan Camgoz and Richard Bowden},
  year          = {2020},
  eprint        = {2004.14874},
  archiveprefix = {arXiv},
  primaryclass  = {cs.CV},
  url           = {https://arxiv.org/abs/2004.14874}
}

@misc{atmc,
  title         = {Adversarial Training for Multi-Channel Sign Language Production},
  author        = {Ben Saunders and Necati Cihan Camgoz and Richard Bowden},
  year          = {2020},
  eprint        = {2008.12405},
  archiveprefix = {arXiv},
  primaryclass  = {cs.CV},
  url           = {https://arxiv.org/abs/2008.12405}
}

@article{Huang2021TowardsFA,
  title={Towards Fast and High-Quality Sign Language Production},
  author={Wencan Huang and Wenwen Pan and Zhou Zhao and Qi Tian},
  journal={Proceedings of the 29th ACM International Conference on Multimedia},
  year={2021},
  url={https://api.semanticscholar.org/CorpusID:239011890}
}

@misc{3dmc,
  title         = {Continuous 3D Multi-Channel Sign Language Production via Progressive Transformers and Mixture Density Networks},
  author        = {Ben Saunders and Necati Cihan Camgoz and Richard Bowden},
  year          = {2021},
  eprint        = {2103.06982},
  archiveprefix = {arXiv},
  primaryclass  = {cs.CV},
  url           = {https://arxiv.org/abs/2103.06982}
}

@misc{ddpm,
      title={Denoising Diffusion Probabilistic Models}, 
      author={Jonathan Ho and Ajay Jain and Pieter Abbeel},
      year={2020},
      eprint={2006.11239},
      archivePrefix={arXiv},
      primaryClass={cs.LG},
      url={https://arxiv.org/abs/2006.11239}, 
}

@misc{signidd,
  title         = {Sign-IDD: Iconicity Disentangled Diffusion for Sign Language Production},
  author        = {Shengeng Tang and Jiayi He and Dan Guo and Yanyan Wei and Feng Li and Richang Hong},
  year          = {2024},
  eprint        = {2412.13609},
  archiveprefix = {arXiv},
  primaryclass  = {cs.CV},
  url           = {https://arxiv.org/abs/2412.13609}
}

@article{gcdm,
  author     = {Tang, Shengeng and Xue, Feng and Wu, Jingjing and Wang, Shuo and Hong, Richang},
  title      = {Gloss-driven Conditional Diffusion Models for Sign Language Production},
  year       = {2025},
  issue_date = {April 2025},
  publisher  = {Association for Computing Machinery},
  address    = {New York, NY, USA},
  volume     = {21},
  number     = {4},
  issn       = {1551-6857},
  url        = {https://doi.org/10.1145/3663572},
  doi        = {10.1145/3663572},
  journal    = {ACM Trans. Multimedia Comput. Commun. Appl.},
  month      = mar,
  articleno  = {105},
  numpages   = {17}
}

@article{g2pddm,
  title   = {G2P-DDM: Generating Sign Pose Sequence from Gloss Sequence with Discrete Diffusion Model},
  author  = {Xie, Pan and Zhang, Qipeng and Taiying, Peng and Tang, Hao and Du, Yao and Li, Zexian},
  journal = {Proceedings of the AAAI Conference on Artificial Intelligence},
  volume  = {38},
  number  = {6},
  pages   = {},
  year    = {2024},
  doi     = {10.1609/aaai.v38i6.28441},
  url     = {https://doi.org/10.1609/aaai.v38i6.28441}
}

@misc{signd2c,
  title         = {Discrete to Continuous: Generating Smooth Transition Poses from Sign Language Observation},
  author        = {Shengeng Tang and Jiayi He and Lechao Cheng and Jingjing Wu and Dan Guo and Richang Hong},
  year          = {2024},
  eprint        = {2411.16810},
  archiveprefix = {arXiv},
  primaryclass  = {cs.CV},
  url           = {https://arxiv.org/abs/2411.16810}
}

@article{slr1,
  author  = {Dan Guo and Wengang Zhou and Houqiang Li and Meng Wang},
  title   = {Online Early-Late Fusion Based on Adaptive HMM for Sign Language Recognition},
  journal = {ACM Transactions on Multimedia Computing, Communications, and Applications},
  volume  = {14},
  number  = {1},
  pages   = {1--18},
  year    = {2017},
  doi     = {10.1145/3152121}
}

@article{slr2,
  author   = {Hu, Hezhen and Pu, Junfu and Zhou, Wengang and Li, Houqiang},
  journal  = {IEEE Transactions on Multimedia},
  title    = {Collaborative Multilingual Continuous Sign Language Recognition: A Unified Framework},
  year     = {2023},
  volume   = {25},
  number   = {},
  pages    = {7559-7570},
  doi      = {10.1109/TMM.2022.3223260}
}

@article{slr3,
  author   = {Hu, Hezhen and Zhao, Weichao and Zhou, Wengang and Li, Houqiang},
  journal  = {IEEE Transactions on Pattern Analysis and Machine Intelligence},
  title    = {SignBERT+: Hand-Model-Aware Self-Supervised Pre-Training for Sign Language Understanding},
  year     = {2023},
  volume   = {45},
  number   = {9},
  pages    = {11221-11239},
  doi      = {10.1109/TPAMI.2023.3269220}
}

@misc{slr4,
  title         = {Quantitative Survey of the State of the Art in Sign Language Recognition},
  author        = {Oscar Koller},
  year          = {2020},
  eprint        = {2008.09918},
  archiveprefix = {arXiv},
  primaryclass  = {cs.CV},
  url           = {https://arxiv.org/abs/2008.09918}
}

@inproceedings{slr5,
  author    = {Weichao Zhao and Hezhen Hu and Wengang Zhou and Jiaxin Shi and Houqiang Li},
  title     = {BEST: BERT Pre-training for Sign Language Recognition with Coupling Tokenization},
  booktitle = {Proceedings of the AAAI Conference on Artificial Intelligence},
  volume    = {37},
  pages     = {3597--3605},
  year      = {2023},
  doi       = {10.1609/aaai.v37i3.25470}
}

@misc{slt1,
  title         = {Multi-channel Transformers for Multi-articulatory Sign Language Translation},
  author        = {Necati Cihan Camgoz and Oscar Koller and Simon Hadfield and Richard Bowden},
  year          = {2020},
  eprint        = {2009.00299},
  archiveprefix = {arXiv},
  primaryclass  = {cs.CV},
  url           = {https://arxiv.org/abs/2009.00299}
}

@misc{slt2,
  title         = {LLMs are Good Sign Language Translators},
  author        = {Jia Gong and Lin Geng Foo and Yixuan He and Hossein Rahmani and Jun Liu},
  year          = {2024},
  eprint        = {2404.00925},
  archiveprefix = {arXiv},
  primaryclass  = {cs.CV},
  url           = {https://arxiv.org/abs/2404.00925}
}

@inproceedings{slt3,
  author    = {Dan Guo and Shengeng Tang and Meng Wang},
  title     = {Connectionist Temporal Modeling of Video and Language: a Joint Model for Translation and Sign Labeling},
  booktitle = {Proceedings of the Twenty-Eighth International Joint Conference on Artificial Intelligence (IJCAI)},
  pages     = {751--757},
  year      = {2019},
  doi       = {10.24963/ijcai.2019/106}
}

@inproceedings{slp1,
  author    = {Seshadri Mazumder and Rudrabha Mukhopadhyay and Vinay P. Namboodiri and C. V. Jawahar},
  title     = {Translating Sign Language Videos to Talking Faces},
  booktitle = {Proceedings of the Twelfth Indian Conference on Computer Vision, Graphics and Image Processing (ICVGIP)},
  year      = {2021},
  articleno = {27},
  numpages  = {10},
  doi       = {10.1145/3490035.3490286}
}

@article{slp2,
  author  = {John McDonald and Rosalee J. Wolfe and Jerry Schnepp and Julie Hochgesang and Diana Gorman Jamrozik and Marie Stumbo and Larwan Berke and Melissa Bialek and Farah Thomas},
  title   = {An Automated Technique for Real-Time Production of Lifelike Animations of American Sign Language},
  journal = {Universal Access in the Information Society},
  volume  = {15},
  number  = {4},
  pages   = {551--566},
  year    = {2016},
  doi     = {10.1007/s10209-015-0407-2}
}

@article{slp3,
  author  = {Segouat, Jérémie},
  year    = {2009},
  month   = {01},
  pages   = {},
  title   = {A Study of Sign Language Coarticulation},
  volume  = {93},
  journal = {ACM Sigaccess Accessibility and Computing},
  doi     = {10.1145/1531930.1531935}
}

@article{text2sign,
  author  = {Stoll, Stephanie and Camgoz, Necati and Hadfield, Simon and Bowden, Richard},
  year    = {2020},
  month   = {04},
  pages   = {},
  title   = {Text2Sign: Towards Sign Language Production Using Neural Machine Translation and Generative Adversarial Networks},
  journal = {International Journal of Computer Vision},
  doi     = {10.1007/s11263-019-01281-2}
}

@article{selfforing,
  title   = {Self Forcing: Bridging the Train-Test Gap in Autoregressive Video Diffusion},
  author  = {Huang, Xun and Li, Zhengqi and He, Guande and Zhou, Mingyuan and Shechtman, Eli},
  journal = {arXiv preprint arXiv:2506.08009},
  year    = {2025}
}

@misc{signdiff,
  title         = {SignDiff: Diffusion Model for American Sign Language Production},
  author        = {Sen Fang and Chunyu Sui and Yanghao Zhou and Xuedong Zhang and Hongbin Zhong and Yapeng Tian and Chen Chen},
  year          = {2025},
  eprint        = {2308.16082},
  archiveprefix = {arXiv},
  primaryclass  = {cs.CV},
  url           = {https://arxiv.org/abs/2308.16082}
}

@misc{flowmatching1,
      title={Flow Matching Guide and Code}, 
      author={Yaron Lipman and Marton Havasi and Peter Holderrieth and Neta Shaul and Matt Le and Brian Karrer and Ricky T. Q. Chen and David Lopez-Paz and Heli Ben-Hamu and Itai Gat},
      year={2024},
      eprint={2412.06264},
      archivePrefix={arXiv},
      primaryClass={cs.LG},
      url={https://arxiv.org/abs/2412.06264}, 
}

@misc{flowmatching2,
      title={Flow Matching for Generative Modeling}, 
      author={Yaron Lipman and Ricky T. Q. Chen and Heli Ben-Hamu and Maximilian Nickel and Matt Le},
      year={2023},
      eprint={2210.02747},
      archivePrefix={arXiv},
      primaryClass={cs.LG},
      url={https://arxiv.org/abs/2210.02747}, 
}

@misc{dit,
      title={Scalable Diffusion Models with Transformers}, 
      author={William Peebles and Saining Xie},
      year={2023},
      eprint={2212.09748},
      archivePrefix={arXiv},
      primaryClass={cs.CV},
      url={https://arxiv.org/abs/2212.09748}, 
}

@inproceedings{gen-obt,
author = {Tang, Shengeng and Hong, Richang and Guo, Dan and Wang, Meng},
title = {Gloss Semantic-Enhanced Network with Online Back-Translation for Sign Language Production},
year = {2022},
isbn = {9781450392037},
publisher = {Association for Computing Machinery},
address = {New York, NY, USA},
url = {https://doi.org/10.1145/3503161.3547830},
doi = {10.1145/3503161.3547830},
booktitle = {Proceedings of the 30th ACM International Conference on Multimedia},
pages = {5630–5638},
numpages = {9},
location = {Lisboa, Portugal},
series = {MM '22}
}

\appendix

\section{Additional Implementation and Evaluation Details}
This appendix provides supplementary details on data preprocessing, training configuration, evaluation protocols, and additional qualitative results.

\section{Datasets and Preprocessing}

\paragraph{Datasets.}
We use two benchmark datasets: PHOENIX14T~\cite{phoenix} and How2Sign~\cite{how2sign}. We follow the official train/validation/test splits for both datasets.

\begin{table}[ht]
\centering
\resizebox{0.49\textwidth}{!}{
\begin{tabular}{lcccc}
\toprule
Dataset & Language & Gloss annotations & Continuous signing & Native pose keypoints \\
\midrule
PHOENIX14T & German & Yes & Yes & No \\
How2Sign & English & Yes & Yes & Yes \\
\bottomrule
\end{tabular}
}
\caption{Overview of the datasets used in this work. ``Native pose keypoints'' indicates whether pose annotations are provided by the original dataset release.}
\label{table:datasets}
\end{table}

\paragraph{Pose extraction and representation.}
PHOENIX14T does not provide pose sequences, whereas How2Sign includes pre-extracted 2D pose data. To keep the representation consistent across datasets, we extract 2D keypoints from PHOENIX14T video frames using OpenPose~\cite{openpose}. The final pose representation contains 137 keypoints per frame, covering the body, hands, and face. Each keypoint is represented by its 2D coordinates together with a confidence score.

\paragraph{Normalization.}
To improve training stability, we normalize the $x$ and $y$ coordinates of all keypoints separately. Confidence scores are already bounded in $[0,1]$ and are therefore kept unchanged. These confidence values are later used by the Confidence-Aware Causal Attention module described in Section~4.4.

\section{Training Details and Hyperparameters}

\paragraph{Hardware and runtime.}
Training is performed separately on the two datasets using a single 80GB NVIDIA A100 GPU. On PHOENIX14T, the model converges in approximately 40 hours; on How2Sign, training requires approximately 45 hours under the same hardware setup.

\paragraph{Optimization setup.}
Unless otherwise stated in the main text, all experiments use the hyperparameters listed in Table~\ref{table:hyperparameter}. The default configuration uses three experts and bfloat16 precision, matching the setting reported for the main results.

\begin{table*}[ht]
\centering
\resizebox{0.9\textwidth}{!}{
\begin{tabular}{lcc}
\toprule
Hyperparameter & PHOENIX14T & How2Sign \\
\midrule
Seed & 42 & 42 \\
Optimizer & AdamW & AdamW \\
Learning rate & 1e-4 & 1e-4 \\
Weight decay & 0.01 & 0.01 \\
Scheduler & ExponentialLR & ExponentialLR \\
Decay factor & 0.5 & 0.1 \\
Batch size & 8 & 16 \\
Epochs & 5 & 3 \\
Dropout & 0.1 & 0.2 \\
Embedding dimension & 512 & 768 \\
Numerical precision & bfloat16 & bfloat16 \\
Number of experts & 3 & 3 \\
\bottomrule
\end{tabular}
}
\caption{Training hyperparameters used in our experiments.}
\label{table:hyperparameter}
\end{table*}

\section{Evaluation Metrics}

Standard text-generation metrics cannot be applied directly to sign pose sequences. Following prior work, we therefore use a pre-trained sign language translation model for back-translation, convert the generated pose sequence back into text, and report both text-based and motion-based metrics. For efficiency reporting, latency denotes time-to-first-frame and throughput denotes the average generated frames per second under the same 60-frame protocol used in the main paper.

\subsection{Text-based Metrics}

\paragraph{BLEU.}
BLEU~\cite{bleu} measures $n$-gram precision between the back-translated sentence and the reference sentence. We report BLEU-1 and BLEU-4 to capture unigram and 4-gram overlap, respectively. Higher BLEU indicates better semantic agreement with the source sentence.

\paragraph{ROUGE-L.}
ROUGE~\cite{rouge} is a recall-oriented metric. We use ROUGE-L, which is based on the longest common subsequence between the back-translated sentence and the reference. Higher ROUGE-L indicates stronger overlap in sentence content and ordering.

\paragraph{WER.}
Word Error Rate (WER) measures the discrepancy between the back-translated sentence and the original reference:
\begin{equation}
\mathrm{WER} = \frac{S + D + I}{N},
\end{equation}
where $S$, $D$, and $I$ denote the numbers of substitutions, deletions, and insertions, and $N$ is the number of reference words. Lower WER indicates better preservation of the input semantics.

\subsection{Motion-based Metrics}

\paragraph{DTW.}
Dynamic Time Warping (DTW) measures the alignment cost between two motion sequences that may differ in temporal dynamics:
\begin{equation}
\mathrm{DTW}(X, Y) = \min_{\pi} \sum_{(i,j)\in\pi} \lVert x_i - y_j \rVert_2,
\end{equation}
where $\pi$ denotes a valid warping path. In our setting, DTW is computed over pose trajectories to evaluate structural and temporal similarity between generated and ground-truth motion. Lower DTW indicates better temporal alignment.

\paragraph{FID.}
Fr\'echet Inception Distance (FID) evaluates the distributional similarity between real and generated samples in a feature space:
\begin{equation}
\mathrm{FID} = \lVert \mu_r - \mu_g \rVert_2^2 + \mathrm{Tr}\left(\Sigma_r + \Sigma_g - 2(\Sigma_r \Sigma_g)^{1/2}\right),
\end{equation}
where $(\mu_r, \Sigma_r)$ and $(\mu_g, \Sigma_g)$ are the means and covariance matrices of real and generated samples, respectively. In our experiments, FID is adapted to pose-based sign generation by extracting features from keypoint sequences. Lower FID indicates more realistic generation.

\section{Additional Qualitative Examples}

\begin{figure*}[htbp]
    \centering
    \includegraphics[width=0.98\linewidth]{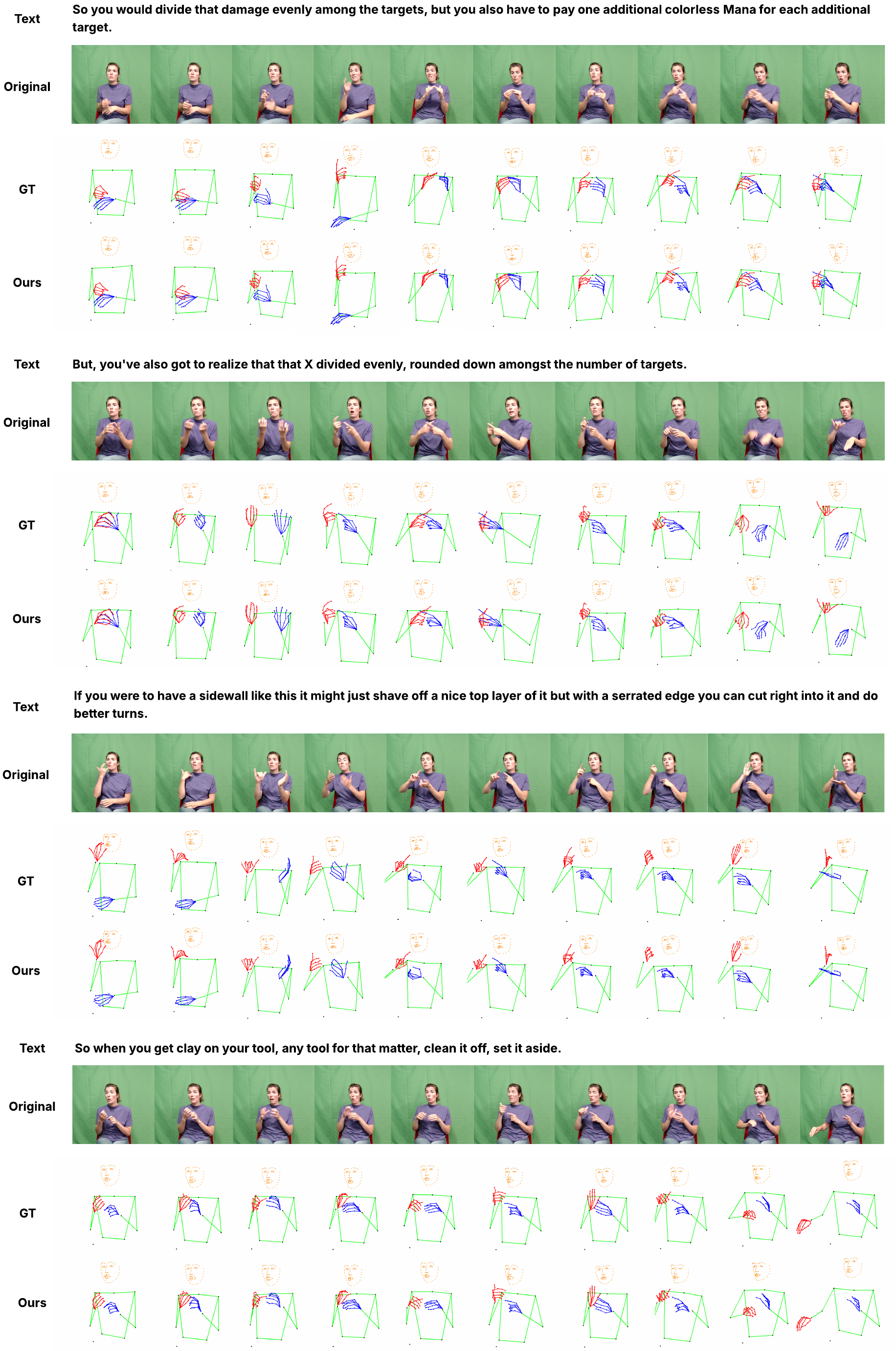}
    \caption{Additional qualitative examples on How2Sign. Compared with Figure~\ref{fig:comparison}, these samples cover more diverse motion patterns and lexical contents, showing that HybridSign remains close to the reference sequence while preserving temporally coherent hand--body coordination.}
    \label{fig:comparison-diverse}
\end{figure*}

Figure~\ref{fig:comparison-diverse} provides additional qualitative examples on How2Sign. Compared with Figure~\ref{fig:comparison} in the main paper, these examples cover more diverse motion patterns and lexical content, further illustrating that HybridSign preserves temporally coherent hand--body coordination while staying close to the reference pose sequence.

\end{document}